\documentclass[letterpaper]{article}
\usepackage{aaai}
\usepackage{times}
\usepackage{helvet}
\usepackage{courier}
\usepackage{mathtools}
\begin{document}
% The file aaai.sty is the style file for AAAI Press 
% proceedings, working notes, and technical reports.
%
\title{DeepCO: Offline Combinatorial Optimization Framework\\
Utilizing Deep Learning}

\author{Wenpeng Wei,\\
{Central Research Lab, Hitachi, Ltd.}\\
wenpeng.wei.bo@hitachi.com,
\And
Toshiko Aizono,\\
{Central Research Lab, Hitachi, Ltd.}\\
toshiko.aizono.jn@hitachi.com
}

\maketitle
\begin{abstract}
\begin{quote}
Combinatorial optimization serves as an essential part in many modern industrial applications. 
A great number of the problems are offline setting due to safety and/or cost issues. 
While simulation-based approaches appear difficult to realise for complicated systems, in this research, we propose DeepCO, an offline combinatorial optimization framework utilizing deep learning. 
We also design an offline variation of Travelling Salesman Problem (TSP) to model warehouse operation sequence optimization problem for evaluation. 
With only limited historical data, novel proposed distribution regularized optimization method outperforms existing baseline method in offline TSP experiment reducing route length by 5.7\% averagely and shows great potential in real world problems. 
\end{quote}
\end{abstract}

\section{Introduction}
With the rapid process of decreasing birth-rate and aging population, labour force shortage becomes serious in many industries world widely. 
As a solution, an increasing number of industrial robots are introduced into fields like manufacturing factories and logistic warehouses.
Robotization eases the urgent lack of labour force on one hand, however on the other hand, it also introduces complexity of control and operation. 
Human workers are now collaborating with robots across multiple processes in the fields. 
Two heterogeneous groups of operators make the planning and controlling of operations extremely difficult. 
The settings and policies of operations which are suitable and optimal to one are possibly not appropriate for the other. Furthermore, there is little experiences and historical data and policies to follow. 
Therefore, to increase productivity in such situation, the approach and technique of optimization are gathering attention from both industrial and research community. 

Optimization serves as an essential part in many modern industrial applications. 
Manufacturing factory optimize producing sequence and operator schedule to improve productivity. 
Retailers optimize assortment and product stock to reduce cost and opportunity loss. 
Optimization aims to find better solutions to problems in business and daily life. 

In Hitachi Group, business based on manufacturing plan optimization and stock optimization is also widely developed as our strength of Operation Technology (OT) in industrial field. 
To achieve that goal, one must either model target problem formally and calculate solution analytically with appropriate algorithms, or improve the solution numerically by modifying and evaluating solution candidate repeatedly against target problem. 
The former approach requires entire knowledge about the target problem, which is usually unrealistic for real world problem. 
Even if nearly entire knowledge is acquired, to keep up with the dynamic open environment which changes rapidly, one needs to update the knowledge regularly.
The maintenance of business knowledge turns into a heavy burden. 
While the later approach, which is called online optimization due to the nature that optimizer requires access to target problem, is not suitable for certain circumstances. 
For the target problems where solution experiment is expensive or dangerous, trying out solution candidates which may lead to unpredictable results is not acceptable.
For instance, changing parameters can probably damage devices in factories, or cause operator injure.
Those type of problems are not able to adopt online optimization approach naively.
One possible work around is to use simulators. 
By optimizing in a simulated environment, cost or safety issue can be avoided. 
Nonetheless, obtaining accurate simulators remains as a difficult problem, especially for complicated systems, where multiple different components interact with each other. 

On the other hand, most of modern industrial applications are becoming digitalized. 
The information of problems and solutions are recorded as historical data.
The movement of utilizing those valuable data with statistical and learning-based approaches makes tremendous success in many areas, such as anomaly detection, demand prediction and recommendation to name a few. 
Considering the situation mentioned above, there emerges both a need and an opportunity for offline optimization based on only historical data without access target problem itself while optimizing.

\section{Related Works}
Combinatorial optimization is well researched for decades in operation research field. For problems those can be formally modelled, researchers propose various algorithms to calculate exact optimal solutions.
Considering the cost and difficulty to model complicated problems, iterative search-based methods are another option.
Coming in the price of relatively long calculation time, search-based methods do not require formally modelling target problem. 

Recently, with the development and success of deep learning, combinatorial optimization through deep learning becomes a rising trend. 
Pointer Network \cite{vinyals2015pointer} firstly utilizes attention mechanism to produce permutation of input sequences. 
Based on it, Kool \cite{kool2018attention} trains neural network to solve routing problems.
The approach is essentially supervised learning, where training data is problem instances with optimal solutions. 
Once the network has been trained, new problem can be solved directly with performance comparable to optimal solutions. 
Clearly, supervised learning approach requires labelled data, which in this case, is optimal solutions to target problems in train set. 
It is a requirement most of industrial applications do not meet.

Another stream of deep learning based combinatorial optimization is the approach based on reinforcement learning (RL).
Similar to traditional search-based methods, RL searches for optimal solution by interacting with the target problem, i.e. evaluating solution candidate, then use the evaluation to improve solution policy. 
It can be seen as a deep learning variation of traditional search-based approach, with learnt searching policies instead of explicitly coded ones. 
The interactive nature of RL requires access to target problem while optimizing, thus raises cost and safety issue mentioned in last section. 
To address that, offline reinforcement learning is considered.
Offline RL learns policy using only limited fixed data, and no evaluation of solution candidate is needed.
Even though the performance of offline RL is not comparable to online ones, it looks promising for complicated optimization problem in offline setting.
However, RL based approaches assume target problem is able to be modelled by Markov Decision Process, which means the state of time t can be decided entirely by the state of time t-1 and action after that. 
Longer period of dependency has to be encoded into state to be captured, which make it extremely difficult to model problem and to train networks. 
Therefore for the problem with long term dependencies, like most of industrial applications, adopting RL based approach is quite difficult and inefficient.

\section{Proposal}

\subsection{Problem Setting}
The problem we are trying to solve is offline combinatorial optimization. 
Offline means the solver has no access to the problem while optimizing to evaluate the performance of a possible solution.
The only available information is a limited number of historical data, each contains the triplet of problem condition, solution and its corresponding performance.
The goal is to find a better solution using only historical data, without acquiring new data by evaluating candidate solutions repeatedly against the problem.

Formally, let F be the problem to be optimized, then
$$ y = F(x,c),   (c,x,y) \in D$$
where $D$ is the historical data, i.e. the training data, and $(c,x,y)$ is condition, solution, and performance respectively. 
Since the problem $F$ is not available to the solver, a surrogate problem $\hat{F}$ is needed to be obtained using learning-based approach, utilizing $D$,
$$ \hat{F} = \arg\min_{f} L(y, f(x,c)),   (c,x,y) \in D$$

where $f$ is hypothesis model, and $L$ is appropriate loss function. 
Then the optimal solution $x^*$ is the solution achieve the best performance evaluated by $\hat{F}$,
$$ x^* = \arg\min_x \hat{F}(x, c) $$

\subsection{Sparse and Imbalanced Coverage of Training Data}
Combinatorial optimization problems usually hold enormous solution spaces, which is the main reason to make it difficult to search.
From the point of view of statistical learning, it is also not easy to learn an estimator to predict well over entire massive discrete space.
Usually it requires sufficient training data and long time for training.
It requires data to be sufficient in two aspects, amount and coverage.
Obviously, the data must have enough amount to train an accurate estimator.
However, the most important point is the coverage of training data.
Training data must be distributed over the entire solution space to train the estimator, otherwise the estimator has no information about the part where no training data exists, thus cannot give usable prediction.
Therefore, the coverage of the training data is extremely important in the learning of surrogate function $\hat{F}$.

Nonetheless, the training data we can obtain is relatively small and biased. 
The training data is essentially the historical data, which are the records of past solutions to the problem. 
For each particular problem condition, the amount of possible solutions is remarkably large, but only one solution is recorded in the training data. 
In other words, the training data is extremely sparse.
Furthermore, the past solutions are possibly not optimal, but follow certain suboptimal policies.
Therefore, the solutions are possibly similar for comparable problem conditions. 
Instead of spreading over entire solution space, training data is distributed crowdedly around the points where suboptimal solutions lay.
The imbalanced nature of training data distribution combined with sparsity makes it incredibly difficult to train an estimator which maintains prediction accuracy while generalizing well to the samples, i.e. problem condition and candidate solution, which do not exist in training data.

\subsection{Overconfident Estimator}
Besides training data, on the other hand, the approach to model estimator also must be considered. 
Although plenty methods exist, in this research we choose Artificial Neural Network (ANN).
It has been proved that ANN can be trained as universal approximator, which states that a feed-forward network with a single hidden layer containing a finite number of neurons can approximate continuous functions on compact subsets of $R^n$, under mild assumptions on the activation function. 
The statement above makes ANN sound for modelling performance estimator. 

Addition to that, ANN is also capable to deal with multiple types of input. 
Both numerical and categorical, continuous and discrete, fixed and variable length input can be processed well accordingly, using different network structure, such as recurrent neural network or attention mechanism. 
The flexibility of input type of ANN is vital for the use of modelling performance estimator, since in real world applications the representation of problem conditions and solutions vary dramatically.  

On the other hand, ANN are prone to be overconfident for data out of distribution of training data. 
In other words, a trained ANN estimator cannot tell the differences between samples it can predict correctly and samples it cannot. 
For the later, it simply gives wrong predictions without warning or knowing it. 
In ordinary machine learning scenario, the samples to be predict and the samples in training data are assumed independent and identically distributed, which means they are all from the same distribution, thus the overconfident prediction for out of distribution samples is not an issue.
However, in the optimization scenario, where the trained estimator is used to evaluate candidate solution, and the training data is sparse and imbalanced distributed, there is high probability that the sample to predict, i.e. problem condition and candidate solution, is coming outside from the training data distribution, makes the prediction barely reliable.
When optimizing against unreliable prediction, the result usually turns bad, obviously. 

\subsection{Distribution Regularized Optimization}
Addressing the problems discussed above, we propose a framework called DeepCO, targeting offline combinatorial optimization problems. 
The core part of DeepCO is a novel method called distribution regularized optimization, which contains mainly three parts as follows.

\subsubsection{Pair-wise ranking for performance estimation}
To train a good estimator through learning-based approach, one important choice of design is the loss function. 
Empirically, squared root error or MSE works well for regression tasks, and cross entropy is considered suitable for classification applications. But in practise, we found that the MSE loss works poorly for training performance regression estimator.
Noticing that optimizing only need to know relative order of the performance instead of absolute value, we propose to train performance estimator as ranking model instead of regression model as follows.

$$ P_{ij} = \frac{1}{1 + \exp{(F(x_i, c_i) - F(x_j, c_j))}} $$

where $P_{ij}$ is the probability that performance of $(x_i, c_i)$ is higher than performance of $(x_j, c_j)$.
By modelling performance estimator as ranking model in this way, limited amount of training data can be used maximally for the fact that each sample can be learnt multiple times in different comparisons.

\subsubsection{Distance from training data distribution}
As discussed above, ANN lacks the ability to tell whether a sample is coming from the training data distribution or not. 
It makes ANN to give overconfident predictions to the out of distribution samples, which prevent optimizing correctly. 
To avoid that, distance from training data distribution must be calculated additionally. 
Many variations of distance definition can be considered, among them mahalanobis distance can be used as an instance. 
Assuming the feature of problem condition and candidate solution $f(x, c)$, i.e. the output of last hidden layer of the network, obeys Gaussian distribution, then the mahalanobis distance can be calculate as follows.

$$ MD(x, c) = (f(x, c) - \mu)^T \Sigma (f(x, c) - \mu)$$

where

$$\mu = \frac{1}{N}\sum f(x_i, c_i)$$

$$\Sigma = \frac{1}{N} \sum (f(x_i, c_i) - \mu) (f(x_i, c_i) - \mu) ^T $$

$$x_i, c_i \in D$$

\subsubsection{Distribution regularized optimization}
With distance from training data distribution calculated, we can finally obtain the last part of our proposed method.
The basic idea is to avoid the optimizer from exploring the area where estimator cannot predict well. 
In the contrary, by forcing the optimizer only searching the part of the solution space where trained predictor is able to predict with high confidence, we can promise the optimization process undertaken correctly.
Formally, let $\pi_\theta$ be the policy to be optimized,

$$ x = \pi_\theta(c) $$ 

then the optimal solution is given when $\theta$ is optimized,

$$ x^* = \pi_{\theta^*} (c) $$

$$\theta^* = \arg \min_\theta \ell (\pi_\theta(c), c) $$

where l is the cost function to be actually optimized through the optimization process, which is designed as

$$ \ell (x, c) = - \hat{F} (x, c) + \lambda \max (0, MD(x, c) - \alpha) $$

where $\hat{F}$ is the estimator we trained with training data, 
$MD$ is the Mahalanobis distance from training data, which is interchangeable with other valid distance definition, 
$\alpha$ is the hyper parameter to control the size of high confident area, 
and $\lambda$ is the hyper-parameter usually setting with extremely large value such as $1e6$.

Intentionally, the cost function proposed above penalize the solution candidate far than $\alpha$ from training data distribution with a large $\lambda$. 
The penalization is so large to become the dominate component of the cost functions, which encourage the optimizer to push the solution candidate moving towards the training data distribution.
Once it enters the $\alpha$ circle,the penalization disappears, and the optimizer is encouraged to reduce the predicted performance then, which is highly reliable in the circle. 
In other words, instead of trying to find the global optimal solution where the estimator has no knowledge and barely confident, proposed cost function inspire the optimizer to search local optimal solution in the area where the estimator predicts correctly. 
As a result, optimization result is expected to be better than the solutions in the training data, yet reliable enough to be adopted in real world applications. 
The intuition of optimization process in shown in Figure 1 as following.

\section{Experiment and Evaluation}
To validate proposed framework, we design an elementary experiment using Travelling Salesman Problem (TSP) with offline setting as follows.
For each problem instance, 100 cities are sampled uniformly between 0 and 1 in 2-dimensional coordination as problem condition. 
A random route is sampled as solution and the length of the route is recorded as according performance.
The triplet of problem condition, solution and performance form one problem instance and the training set contains 1000 instance.
For test set we sample 200 instances, each contains from 40 to 120 cities.
Noticing the number of cities in test set is variable and different form train set to evaluate how well proposed framework generalize to problems with fundamentally different and more complicated setting.

We train a performance estimator as proposed using training data set mentioned above, then apply simulated annealing method to optimize solution, i.e. the permutation of route to travelling through all cities in the problem condition, with respect to the cost function proposed last section, using only trained estimator to evaluate solution candidate instead of calculate actual length of the route.

We also optimize solution with simulated annealing method with same parameters with respect to estimator predicted performance directly instead of proposed cost function as baseline. 
The evaluation metric is how much better the optimal solution coming from proposed method comparing to baseline, assuming the baseline performance of according problem instance as 1 for normalization. 
The result is shown in the following table.

\begin{center}
\begin{tabular}{ c | c }
  Number of cities & Route length reduction  \\ 
  \hline
  \hline
  $N < 60$ & 3.5\%  \\
  \hline
  $60 \le N < 80$  & 4.1\%  \\
  \hline
  $80 \le N < 100$  & 5.3\%  \\
  \hline
  $100 \le N < 120$  & 6.1\%  \\
  \hline
\end{tabular}
\end{center}

From the table we can observe that the more complicated the problem is, the more route length is reduced comparing to baseline method. 
What is noteworthy is proposed method shows strong performance for the problems with more cities ($100 \le N < 120$ ) than the ones in training data (N = 100). 
This result shows proposed method can generalize to more complicated problem than training data.

Besides the overall performance of proposed framework, we also investigate the optimizing process of both proposed method and baseline. 
By calculating the actual performance of each solution candidate through optimization, we can see clearly the differences between the two method.

% TODO: insert fig 2

As shown in Figure 2, proposed method effectively stops the rebound of optimized result in the contrary to baseline method. 
The rebound is due to the miss prediction of estimator. 
This result shows proposed method effectively regularized the search of solution inside the area where estimator can predict correctly.

\section{Conclusion}
Combinatorial optimization serves as an essential part in many modern industrial applications. 
A great number of the problems are offline setting due to safety and/or cost issues. 
While simulation-based approaches appear difficult to realise for complicated systems, we propose DeepCO, a learning-based framework for offline combinatorial optimization problem, which learns estimator using only limited historical data without access to target problem while optimizing. 
The novel proposed distribution regularized optimization method is shown to outperform existing baseline method in offline TSP experiment reducing route length by 5.7\% averagely.

Combinatorial optimization covers various problems. 
Proposed method is only feasible for the problem where the mapping from solution to evaluation is possible be approximated as continuous function.
Formally,let $F(x)$ be the evaluation of solution $x$, and $D(x_1, x_2)$ be the distance between solution $x_1, x_2$ then proposed method is feasible if and only if

$$ |F(x_1) - F(x_2)| \le K |D(x_1, x_2)| $$

where $K$ is a constant value. 
This nature guarantees that evaluations of similar solutions do not differ much. 
Thus, makes neighbour of solutions meaningful and predictable, which is the fundamental requirement of proposed method and most of machine learning approaches. 
It is necessary to investigate the range of $K$ where proposed method is feasible.

The urgent next step is to evaluate proposed framework in real world problem. 
Also, adopting more sophisticated optimization method rather than simple simulated annealing can possibly lead to further performance improvement. 
Finally, expansion to other type of optimization problems is needed for various business considerations.

\bibliographystyle{aaai} \bibliography{ref}

\end{document}